\documentclass[a4paper,10pt, twocolumn]{article}

\usepackage{amsmath}
\usepackage{amsfonts}
\usepackage{amssymb}
\usepackage{a4wide}
\usepackage{bm}
\usepackage{graphicx}
\graphicspath{ {./images/} }
\usepackage{hyperref}
\usepackage{url}
\usepackage{collectbox}

\DeclareMathOperator*{\argmin}{argmin}
\DeclareMathOperator*{\mostcommon}{most\,\,\,common}

\begin{document}

\title{\bf Towards conceptual generalization in the embedding space} 
                        \vskip25pt
\author{\bf Luka Nenadovi\'c$^{1} \phantom{\,}^{2}$\footnote{Author names are given in alphabetical order.}, \ 
     Vladimir Prelovac$^{1} \phantom{\,}^{2}$\\
                  \\$\strut^{1}$\url{{lnenadovic,vprelovac}@kagi.ai}
                   \\[15pt]$\strut^{2}$Kagi.ai Research\\
                   \\
       }
\maketitle
\parskip 10pt plus2pt minus2pt
\abstract{Humans are able to conceive physical reality by jointly learning different facets thereof. To every pair of notions related to a perceived reality may correspond a mutual relation, which is a notion on its own, but one-level higher. Thus, we may have a description of perceived reality on at least two levels and the translation map between them is in general, due to their different content corpus, one-to-many. Following success of the unsupervised neural machine translation models, which are essentially one-to-one mappings trained separately on monolingual corpora, we examine further capabilities of the unsupervised deep learning methods used there and apply some of these methods to sets of notions of different level and measure. Using the graph and word embedding-like techniques, we build one-to-many map without parallel data in order to establish a unified vector representation of the outer world by combining notions of different kind into a unique conceptual framework. Due to their latent similarity, by aligning the two embedding spaces in purely unsupervised way, one obtains a geometric relation between objects of cognition on the two levels, making it possible to express a natural knowledge using one description in the context of the other.}

\thispagestyle{empty} 

\section{Introduction}

It has been shown recently that it is possible to build a bilingual dictionary for a pair of languages in an unsupervised way \cite{Conneau2018WordTW, artetxe-etal-2017-learning} with very high accuracy. These models even outperform their supervised counterparts on certain tasks. Building neural translator based on this approach proved to be successful \cite{lample2018unsupervised,lample-etal-2018-phrase}. 

The mentioned models are based on the distributional similarity between the word embeddings, i.e. representations of words in finite-dimensional linear vector space, with usually up to $300$, and sometimes even more dimensions. The best-known word embedding model is Word2Vec, made by Mikolov et al.\ \cite{DBLP:journals/corr/MikolovSCCD13}. It is constructed by taking into account statistical measure of the co-occurrence of words in a given text corpus. The words which have similar meaning are represented by vectors which are closer in higher-dimensional linear vector space and vice versa. It turned out soon that there was similarity between distributions of words in a word embedding for different languages, even for distant families of languages \cite{DBLP:journals/corr/MikolovLS13}. Following this, Conneau et al. \cite{Conneau2018WordTW} successfully learned how to relate the word embedding distributions for language pairs, enabling the same words in different languages to match approximately in a common embedding space. The translation consists roughly in identifying them as nearest neighbours. This method is grosso modo a combination of the adversarial training, which approximately aligns the distributions in the embedding space, and the linear transformation called orthogonal Procrustes which then refines this mapping. 

Described method suffices for alignment of two sets of embedded vectors, which have approximately the same measure so that, up to the synonyms, the mapping between different languages is almost one-to-one. This is a consequence of the semantic similarity between  the language corpora in their entirety. Following just mentioned approach we would like to examine the possibility of simultaneous learning and aligning semantically similar structures which describe the same physical system on two different levels, using two sets of data, but with the different measure. The resulting correspondence will not be one-to-one, as in the case of languages, but rather many-to-one (or one-to-many). Our aim is to mimic a mental representation of a set of objects and a set of meta-objects which are of different size and to relate them into unique semantically meaningful cognitive picture. To every two physical states which are causally connected, there is one transition law which enables transition between the states and describes this relation. Thus, we conjecture a similarity of the embedding spaces for the states and transition laws, which will, after we align the distributions of the two vector representations in a common embedding space, reveal this relational structure. This should augment the general knowledge about the given physical system. The agent will learn by analogy to relate two unrelated sets of notions into a unique notional picture. 

There is a plethora of the two-level systems of interest for task-specific artificial intelligence. Firstly, we can combine different monolingual embedding spaces and try to align word embeddings with phrase embeddings or sentence embeddings \cite{cer-etal-2018-universal}. Here, every sentence is related to several words. So far, one can combine pictures with texts, jointly learn visual and verbal content \cite{DBLP:journals/corr/abs-1904-12584}, caption images, or even generate one from the other \cite{Reed:2016:GAT:3045390.3045503}, where to every picture correspond several other textual or notional inputs. All those tasks are specific. We are interested here primarily in notional unification on the most general and primitive level, based solely on statistics, where all learning is inductive, starting from tabula rasa. In this way we want to imitate a very early developmental phase of the human and build a learning agent capable of understanding the basic relations, independent on their content.

In order to generate an outer world which could have two possible and equivalent descriptions we will use the graph model of physical system. This will result essentially in graph embedding. So far, there is an increasing interest in exploring the graph embeddings based on analogy with Word2Vec or similar methods \cite{GOYAL201878}. The resulting vector spaces may be equipped with the rich algebraic structure which then enables different manipulation which is easier and at lower computational cost in several cases. On the other hand, vector embedding is lower-dimensional dense representation, compared to adjacency matrices in case of graphs (or one-hot-encoding in case of words). Our method resembles DeepWalk \cite{Perozzi:2014:DOL:2623330.2623732} and Node2Vec \cite{Grover:2016:NSF:2939672.2939754}, with the difference that we have a fixed graph and do not really use random walk, although there is some randomness in certain transitions. Additionally, we use both node and edge embeddings (Sec.\ \ref{sec3}). 

The plan of this paper is following.\footnote{The source code for this paper is available at \url{github.com/kagi-ai/concept-unification}}  First, in section \ref{sec2} we will build a model which represents a physical system that can be described at two levels. One level is a pure sequence of physical states appearing as system evolves, and the other is a sequence of relations between physical states which plays a role of physical laws of transition applied in the course of the evolution of the system. We will construct this model in such a way that 
\begin{equation*}
\parbox[c]{6em}{\, Kinematical\!\!\\ \phantom{aa!}notions\,}\phantom{!} \longleftrightarrow\phantom{aa} \parbox[c]{5em}{\, Dynamical\\  \phantom{a!}notions\,}
\end{equation*}
relation is not one-to-one, but rather one-to-many. To every physical state correspond more state transition laws. Kinematics of the system will be described in terms of sequence of physical states, whereas the dynamics as the sequence of transition laws. We will further in section \ref{sec3} try to attach the meaning to objects of the system by constructing the embedding vector space using Word2Vec model. This should mirror the semantic structure of the sequences obtained during the evolution, based  solely on the co-occurrence of the states and laws. As a consequence, it should contain the information on the background causal structure of the system. The possibility of description of a physical system is thus twofold and we will build two embedding vector representations independently for each level of description. Further, in section \ref{sec4}, by relating the two vector representations, we will build a unified vector representation of the system. This should in geometrical terms give semantic relation between these two-levels of description. To this end, we will investigate the capabilities of the mentioned orthogonal Procrustes method to build sensible common embedding space for the representation of physical states and transition laws. After that in section \ref{sec6}, we present quantitative analysis of the model. In  the section \ref{sec5}, we discuss our results and outlooks.

\section{Graph representation of a physical system}\label{sec2}

In order to test our conjecture we will investigate a following setup. Our learning agent learns about some outer system of objects from physical world by perceiving their physical states. We will call an ensemble of possible physical states a system of physical states, $\mathcal S$. The system allows for transition between physical states according to some rules. Henceforth, we will call the set of these possible transitions the set of transition laws. The states $\sigma$ of the system $\mathcal S$ are denoted by ordinary numbers $\sigma = 1, 2, 3,\ldots, n$, while the transitions $\tau$ between the states are denoted by the number of first and second state with a letter $s$ between the numbers. This notation is chosen in order to make easier keeping track of relational structure of this two-level system. For example, the transition of the state $\sigma_1$ to the state $\sigma_2$ is denoted by $\sigma_1 s\,\sigma_2$. The set of all transition laws will be denoted as $\mathcal L(\mathcal S)$.

\begin{figure*}[h]
\begin{center}
\includegraphics[scale=0.65]{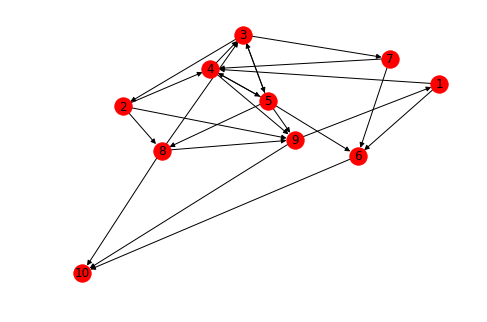}
\caption{Graph $\mathcal{G(\mathcal S)}$, represents a system of physical states $\sigma\in\mathcal S$ and corresponding state transition laws $\mathcal L(\mathcal S)$. Nodes of the graph are physical states $1, \ldots, n \in \mathcal S$ and arrows represent transition laws.  In this specific  example, the number of the nodes is $n = 10$. If our learning agent perceives the state $1$, she can go to the next states $4$ and $6$, while she cannot reach directly the state $10$, since nodes $1$ and $10$ are not directly connected. This can be described using transition laws $1s4$ and $1s6$. Notation is not commutative since the graph is oriented and from this graph it is clear that the transition $9s1$ exists, while $1s9$ is not possible. Due to a present noise, all these transitions are with equal probability. Since there are two directions of transition $4$ to $5$, there is a possibility of a loop $4 \longleftrightarrow 5$.}\label{graph1}
\end{center}
\end{figure*}

The evolution of the physical system can be represented using an oriented graph. In that case, the transition from the state $\sigma_1$ to the state $\sigma_2$ has the form
\begin{equation}
\sigma_1\longrightarrow \sigma_2.
\end{equation}
We will also allow for bidirectional transitions of the type
\begin{equation}
\sigma_1 \longleftrightarrow \sigma_2,
\end{equation}
which could happen in case of an isolated system spontaneously exchanging two states, as a consequence of some conservation law.  In the expanded form, this can be denoted as
\begin{equation}
\sigma_1 \longrightarrow \sigma_2 \longrightarrow \sigma_1 \longrightarrow \sigma_2 \longrightarrow \cdots
\end{equation}
Furthermore, we will enable some noise in the system such that it is possible for a state to go into several different states if it is allowed by the given transition laws. In our analysis, we use the simplest possible case, equal probabilities for transition if the transition laws allow it. For example, if there are allowed transitions
\begin{equation}
\sigma_1\longrightarrow \sigma_2, \qquad \sigma_1\longrightarrow \sigma_3, \qquad \sigma_1\longrightarrow \sigma_4,
\end{equation}
all those transitions are possible with equal probability. Using this, we construct a randomly generated graph $\mathcal{G(\mathcal S)}$\footnote{We created this graph using NetworkX API, \cite{SciPyProceedings_11}.}, which represents a simple system of physical states and the corresponding transition laws, (Fig. \ref{graph1}). 
Our final goal is to allow our learning agent to form latent mental representation about her interaction with the physical system $\mathcal S$. To probe the system we will let her interact with a randomly chosen state $\sigma_i \in \mathcal S$ and follow the evolution of states of the system, which results in a sequences of states. For example, for a given graph
\begin{align}
& 2, 8, 6,\nonumber\\
& 6, 3, 4, 9, 10,\label{stateq}\\
& 1, 3, 2, 10,\nonumber\\
& \cdots\nonumber
\end{align}
where she randomly chooses the length of a sequence, with maximal length of $20$, for the analyzed graph. The corresponding laws are
\begin{align}
& 2s8, 8s6,\nonumber\\
& 6s3, 3s4, 4s9, 9s10,\label{seqeq}\\
& 1s3, 3s2, 2s10,\nonumber\\
&\cdots\nonumber
\end{align}
One may notice that the second and the third sequence here terminate in the state $10$, which in this case plays a role of an equilibrium state. When the system enters this state it stays there for a long time, because there is no way out. 

Edges here describe the higher level notions. For instance, if a state $\sigma_1$ represented by node corresponds to a cold temperature in the room, while a state $\sigma_2$ corresponds to a warm temperature, then the edge $\tau_1=\sigma_1s\,\sigma_2$ can be related to heating the room and $\tau_2=\sigma_2s\,\sigma_1$ to cooling, and so on. We may say that temperature in the room is cold and then warm, or we may say that the temperature in the room rises, which is higher-level, relational description. In this way we established the two descriptions of a given environment and its evolution, each of which requires different number of notions.
We assert here that these two ways of describing are equivalent, which in this case means that we need more higher-level notions. At the same time,
using higher-level description results in shorter sequences.

The sequence of physical states obtained by following this oriented graph is pure kinematic description. On the other hand, the oriented edges of the graph, which represent transition laws, are consequence of the underlying physical laws. Therefore, we can interpret the sequences of transition laws as part of the dynamics of the system. They can be viewed as either different physical laws, or projection of one physical law onto the pair (an element of the product space) of the corresponding physical states. The dynamics of the system is here a higher-order description than the kinematics. On the other hand, if the physical states are just positions, the transitions could correspond to velocities, which are higher level kinematical notions; if the states are points in the phase space, transition laws might represent acceleration, which on the classical level might correspond to a force or another dynamical quantity.

\section{Vector representation of states and laws}\label{sec3}

So far, there are several methods developed in natural language processing (NLP) which can learn geometric representation of words, where the meaning of the words follows solely from their co-occurrence. The models are all based on distributional hypothesis of Harris \cite{harris}, which states that all words that have similar meaning appear in similar contexts. This reasoning can be extended in our case to the graph based physical system $\mathcal{G}(\mathcal S)$, where a learning subject establishes a causal relation between physical states that occur in similar contexts. The treatment here will be exclusively statistical, since we want to investigate capability of an agent to learn about outer world in purely unsupervised way. An agent should build a latent vector representation of physical states. The states are not characterized by any other quality but their number. They are given as pure objects of cognition.

Representing words as vectors which form continuous linear vector space is relatively old idea \cite{Hinton:1986:DR:104279.104287}, but it gained momentum with the deep learning based CBOW and skip-gram Word2Vec model by Mikolov et al. \cite{DBLP:journals/corr/MikolovSCCD13}. This model is purely unsupervised, since it does not need labeled data. By applying this model on a text corpus, one obtains a vector representation which in certain sense reflects semantically relevant properties of the words. The resulting distribution is a latent vector representation which comes from the hidden layer of a neural network of Word2Vec model and at the first glance might not have a direct meaning to us. Nevertheless, it turns out that proximity of states reflects some semantic (causal) relations between the states. Vectors representing words which are causally similar are close in the resulting vector space and vice versa.
The very idea of giving the geometric form to some semantic set is powerful since it allows algebraic treatment of semantic objects. Moreover, recent research suggests that it has some roots in physiology \cite{Bellmundeaat6766}.

Beside Word2Vec there are several other newer and in certain sense more advanced methods to make word embedding like GloVe, \cite{Pennington14glove:global} and FastText \cite{bojanowski2017enriching, joulin2017bag}. Nevertheless, we do not need to take into account the internal or syntactical structure of words to learn word representation, which can be done with FastText, for instance. Instead, we have here a primitive or pure cognition of a physical system, defined in an abstract way, which could correspond more to the case of pre-lingual knowledge, thereby Word2Vec is sufficient for our needs. 

Henceforth, except otherwise stated, everything we discuss will refer to the graph example from the previous section. Using skip-gram Word2Vec method with the windows size of $5$ from the Gensim library \cite{rehurek_lrec}, we build two separate embedding vector spaces using sequences of physical states (\ref{stateq}) and sequences of physical laws (\ref{seqeq}). We train our model using $10000$ randomly initialized sequences in both cases. Here we choose a relatively small number of dimensions in order to present our results graphically. For visualization we use principal component analysis (PCA), a statistically based dimensionality reduction technique \cite{Tipping99probabilisticprincipal}, from Scikit-learn \cite{scikit-learn}. It can distort some relations when projecting from higher-dimensional space when the number of dimensions is too high. Our choice for the number of embedding dimensions therefore is relatively low, $d=10$.

\begin{figure*}[h]
\begin{center}
\includegraphics[scale=0.33]{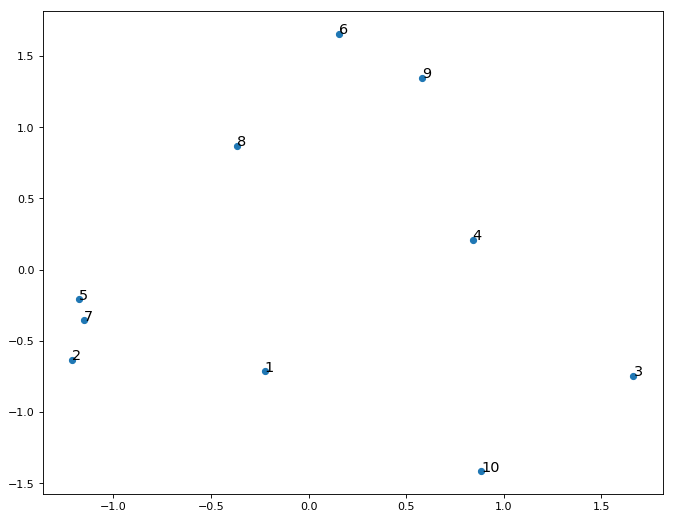}
\includegraphics[scale=0.33]{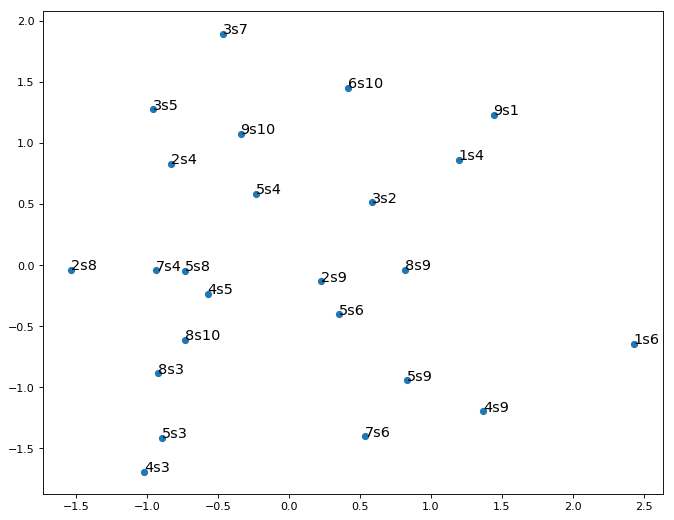}
\caption{PCA projection of the two independently learned embedding vectors. The vector embedding representation $\mathcal{E}(\mathcal S)$ for physical states (left) mirrors certain contextual similarities between physical states. Similarly for the transition laws embedding $\mathcal{E}(\mathcal L)$ (right).}\label{graph2}
\end{center}
\end{figure*}

After applying Word2Vec on sequences of states and laws, we obtain two embedding spaces. The embedding vectors for states $\sigma$, are $d$-dimensional vectors $\bm\sigma$ and they belong to the embedding vector space $\mathcal{E}(\mathcal S)$. There are $n=10$ such vectors and we can collect them into a single matrix $\bm X$ of the dimension $d\times n$. Similarly, the vector encoding for $m=21$ transition laws results in $d\times m$ matrix $\bm Y$, where we collected vectors $\bm\tau$ for the embedding of transition laws $\mathcal{E}(\mathcal L)$. From now on, we will refer to $\bm X\in \mathcal{E}(\mathcal S)$ and $\bm Y\in \mathcal{E}(\mathcal L)$ as to embedding vectors. The resulting $10$-dimensional embedding spaces have some distribution of vectors which reflects latent semantic structure, like similarities and dissimilarities between the corresponding physical states and transition laws (Fig. \ref{graph2}).
In this way we represented our synthetic example of a physical world, using two levels of description with different measure, which was our aim. With this choice of density of graph edges pro node, we have $\varrho = 2.1$ times more transition laws than states. This can be viewed as interaction of agent with a system where she perceives the sequences of states she experiences and then goes to read the textbook which describes the evolution of the given physical system in terms of corresponding transition laws. We want to examine her capability to form a unique mental representation which could comprise both the instances of her direct experience of physical states and relational mental objects expressed in terms of transition laws.

This two-level description is based on the word embedding techniques, but it can also be related to some generic embedding. Especially if the laws can be stated in terms of sentences or phrases, some very advanced embedding models can be taken into account, like the universal sentence encoder \cite{cer-etal-2018-universal}, where the whole sentences are embedded in a vector space of a very high dimensionality. In this case the position of a sentence reflects latent semantic property of a sentence. 

\section{Building a unique representation space}\label{sec4}

In this section we want to build a unique latent vector representation for the set of physical states and related transition laws. We will refer to it as to the mental representation of a learning agent, due to the existence of underlying semantic structure and causality. We conjectured that aligning distributions of the vectors in the embedding representation space built independently for the states $\mathcal{E}(\mathcal{S})$ and laws $\mathcal{E}(\mathcal{L})$ should result in some sensible result, since it describes two aspects of the same physical reality. 

\begin{figure*}[h]
\begin{center}
\includegraphics[scale=0.60]{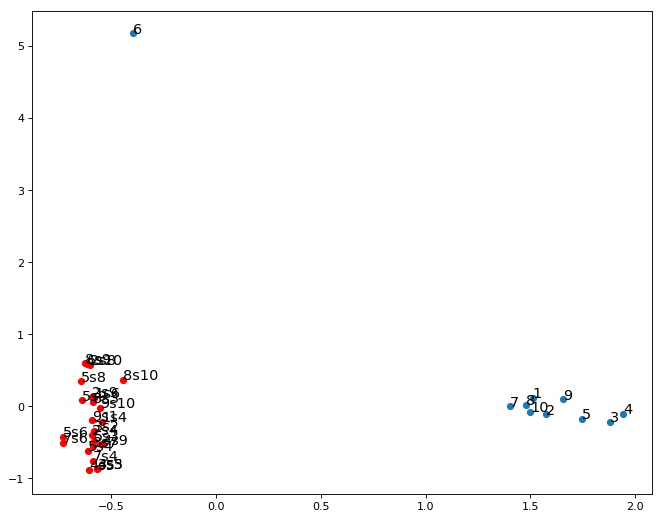}
\caption{PCA projection of common embedding space for states and transition laws before orthogonal Procrustes.}\label{graph3}
\end{center}
\end{figure*}

When we put together these independently trained embedding representations in a common $10$-dimensional space we obtain two separated and non-related distributions like in Fig.\ \ref{graph3}. This result is expected since the two sets of sequences, one for physical states and the other for transition laws have fully different statistics. Yet, they should posses certain latent similarity properties.
One way to relate the distribution of obtained embedding vectors of physical laws $\bm Y\in\mathcal{E}(\mathcal{L})$ to the vectors of physical states $\bm X\in \mathcal{E}(\mathcal{S})$  in a common embedding space is called orthogonal Procrustes algorithm. We follow the approach of \cite{RePEc:spr:psycho:v:31:y:1966:i:1:p:1-10, kementchedjhieva-etal-2018-generalizing} where this method is elaborated and used. The point is to apply linear mapping $\Omega: \mathcal{E}(\mathcal{S})\rightarrow \mathcal{E}(\mathcal{L})$, where $\Omega$ is orthogonal, i.e., $\Omega^{T}\Omega = I$, which minimizes the difference in Frobenius norm of the mapping
\begin{equation}
\omega = \argmin_{\Omega}||\bm X -\Omega \bm Y||.\label{procruste}
\end{equation}
The applied orthogonal mapping $\Omega$ acts on the vectors from $\bm Y\in\mathcal{E}(\mathcal{L})$ by performing rotations and reflections. The minimization of the norm in the equation (\ref{procruste}) results in $\Omega$ which makes the mutual distance between the whole distributions in $10$-dimensional space minimal. 
This problem has solution in terms of singular value decomposition (SVD) of the form
\begin{equation}
\omega = UV^{T},\label{proc}
\end{equation}
where the SVD of the matrix $\bm X\bm Y^{T}$ can be written in the form 
\begin{equation}
\bm X\bm Y^{T}=U\Sigma V^T.    \label{SVD}
\end{equation}
Here, the orthogonal matrices $U$ and $V$ correspond to rotations and reflections, while $\Sigma$, which is a diagonal matrix of eigenvalues of $\bm X\bm Y^{T}$, corresponds to scaling transformations. Thus, from the equation (\ref{proc}), it follows that this transformation consists of rotations and reflections.
For the orthogonal Procrustes algorithm (\ref{procruste}) to work $\bm X$ and $\bm Y$ should be of the same shape. Since there are more laws than states ($n$ states and $m$ laws), we have to somehow restrict the number of transition laws taken into account. Our choice is to restrict the vector $\bm Y$ to only the $n$ most commonly applied transition laws before trying to align the two distributions. The restricted $\bm Y$ vector is then
\begin{equation}
\bm{\tilde{Y}}_{d\times n} = \mostcommon_{n\text{\,\,\,of\,\,\,}m}\{\bm Y_{d\times m}\}.
\end{equation}
Following this restriction, the usual SVD can be applied in order to learn the $d\times d$ matrix $\tilde \Omega$ which maps $\bm{\tilde Y}$ to $\bm X$. Thereafter, we apply the matrix $\tilde \Omega$ to the whole vector $\bm Y$ in order to align the distribution of laws from vector $\bm Y$ to the distribution of states, represented by the vector $\bm X$
\begin{equation}
\bm \Upsilon=\tilde \Omega \bm Y.  \label{tran}  
\end{equation}
Together, vectors $\bm X$ and $\bm \Upsilon$ belong to the common $10$-dimensional embedding space $\mathcal E(\mathcal C)$.

\begin{figure*}[h]
\begin{center}
\includegraphics[scale=0.60]{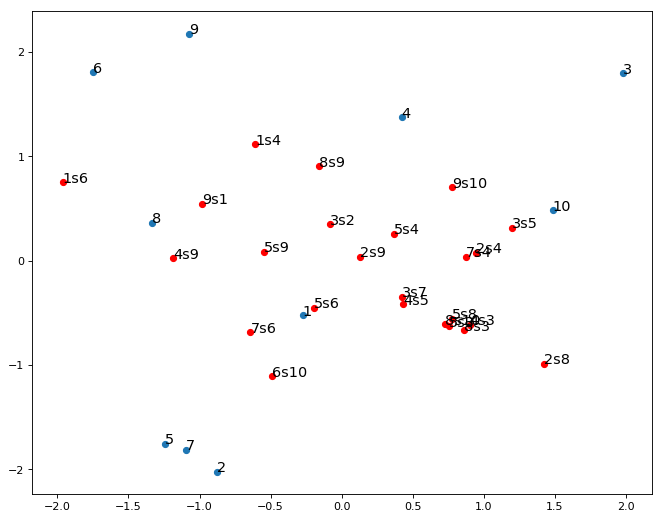}
\caption{PCA projection  after applying orthogonal Procrustes on the set of transition laws. We obtain a unified interrelated representation of physical states (blue) and physical laws (red). Transition laws  in certain cases interpolate between the corresponding physical states.}\label{graph4}
\end{center}
\end{figure*}

The obtained result\footnote{We used orthogonal Procrustes from SciPy, \cite{scipy}.} in the common embedding space is visualized using PCA dimensional reduction in the Fig.\ \ref{graph4}.
The result suggests that there are some regularities in the geometric picture we obtained. Certain laws are placed somewhere between the corresponding states, belonging to the neighbours of the middle-point between the corresponding states in this $10$-dimensional space. 
Resulting alignment of the two distributions allows us to see clearly that the law $3s2$ is placed near the half-way between the states $3$ and $2$. Similarly holds for $9s1$, $3s5$, $3s7$, $9s10$, etc. 
This regularity may be expressed in the form
\begin{equation}
\bm \tau \approx  \frac{1}{2}(\bm \sigma_1+\bm \sigma_2),\label{triangle1}
\end{equation}
where $\bm \tau$ is the vector representation of the transition law $\tau=\sigma_1 s\, \sigma_2$ and $\bm \sigma_1$ and $\bm \sigma_2$ are vector representations of the corresponding states. The approximation sign correspond to the the vector at minimal distance, that is the nearest neighbour. This is expressed in Euclidean metric, but may be as well stated in terms of cosine distance, in which case the nearest neighbour correspond to maximum of the projection 
\begin{equation}
\frac{\bm \tau \cdot (\bm \sigma_1+\bm \sigma_2)}{|\bm \tau||\bm \sigma_1+\bm \sigma_2|},
\end{equation}
where we have a usual Euclidean dot product and the Euclidean $L_1$ norm. In most cases the vectors of transition laws are distorted from the nearest neighbours of a geometric middle-point for the corresponding states, making similar triangles. In these cases there are approximate analogies, for instance
\begin{align*}
&5, 6\leftrightarrow 5s6 \quad \sim  \quad 7, 6,\leftrightarrow 7s6, \\
&5, 9\leftrightarrow 5s9 \quad \sim  \quad 2, 9,\leftrightarrow 2s9, \\
&5, 4\leftrightarrow 5s4 \quad \sim  \quad 7, 4,\leftrightarrow 7s4,
\end{align*}
which can be expressed as
\begin{equation}
\bm \tau_1- \frac{1}{2}(\bm\sigma+\bm\sigma_1)\approx\bm\tau_2- \frac{1}{2}(\bm\sigma+\bm\sigma_2),\label{triangle3}
\end{equation}
where $\tau_1=\sigma s\, \sigma_1$ and $\tau_2=\sigma s\, \sigma_2$. This suggests us that there is additional alignment needed in order to obtain full interpolation.  The deviation from the interpolation equation (\ref{triangle1}) suggests that orthogonal Procrustes is maybe is not enough to fully prove our conjecture. Nevertheless, this is enough strong sign of distributional similarity between the two embeddings, since the regularities defined by the equations (\ref{triangle1}, \ref{triangle3}) are present in nearly $60\%$ of transition laws in this simple example. 

The two regularities can be expressed as more general statement that vector of transition law $\bm\tau$, where $\tau = \sigma_1 s\, \sigma_2$,  and the corresponding states vectors $\bm\sigma_1$ and $\bm\sigma_2$ form isosceles triangles, or
\begin{equation}
|\bm\tau - \bm\sigma_1|\approx |\bm\tau-\bm\sigma_2|.\label{triangle4}
\end{equation}
This means that the vector of transition laws populate regions between the corresponding states preserving the triangular structural relation with the states. 

Additionally, we can see that transition laws $4s5$ and $5s4$ are near neighbours and that they are placed somewhere between the states $4$ and $5$. The distinction between the given two-directional laws interestingly defines a latent ordering. We will comment the ordering problem in the discussion section.

Contrary to the obtained distribution alignments, if we just shrink  random rotate and translate the whole distribution from Fig.\ \ref{graph3} by hand so that the centroids of the  distributions match, we do not obtain any sensible result (Table \ref{tabula2}). 

\section{Experiments}\label{sec6}

In order to support our claims from the previous section, we apply the equation (\ref{triangle4}) in such a way that we keep first $\bm\tau$,  $\bm\sigma_1$ and $\bm\sigma_2$, and iterate through all transition laws vectors $\bm\upsilon\in \bm \Upsilon$ in $d = 10$ dimensional space in order to find the corresponding nearest neighbour according to 
\begin{equation}
|\bm\tau - \bm\sigma_1|\approx |\bm\upsilon-\bm\sigma_2|.\label{triangle4}
\end{equation}
The embedding vectors are normalized and we use Euclidean $L_1$ norm for the results that follow. 
The statistics we apply here is the following: Top 10\% is the percent of all cases when the $\bm\upsilon$ is among 10\% nearest neighbours for given $\bm \tau$. Similarly, we use Top 30\% and Top 50\%. For all iterations of $\bm\upsilon\in \bm\Upsilon$ we also take the average placement of the right answer Avg, in the set of all iterations. We use percents rather than actual average placement due to the different choices of the number of states and transition laws. This means that if  Avg is 30\%, the correct answer is on average at the 30th place as the nearest neighbour in the case when there are 100 transition laws.

For the example from the previous section with 10 states and 21 corresponding transition laws, i. e. density of $\varrho=2.1$ laws pro state, we randomly generated $1000$ graphs. The obtained statistics shows pretty good stability with standard deviations
$\sigma_{\text{Top} 10\%} = 5\%$, $\sigma_{\text{Top} 30\%} = 8\%$, $\sigma_{\text{Top} 50\%} = 15\%$ and $\sigma_{\text{Avg}} = 5\%$. Results are given in the Table \ref{tabula1}.
\begin{table}
\begin{center}
\begin{tabular}{ c | c | c | c  }
\hline
  Top 10\% & Top 30\% & Top 50\%  & Avg \\
  \hline
  31 &   52  & 80 & 29
\end{tabular}
\caption{Scores in percent for transition laws after orthogonal Procrustes.}\label{tabula1}
\end{center}
\end{table}
On the contrary, if we shuffle the transition laws by randomly swapping their vectors, we get the results as in the Table \ref{tabula2}, which shows the absence of any regularity. In ideal case, the statistics for complete randomness of distribution is:  
Top 10\% = 10\%, Top 30\% = 30\%, Top 50\% = 50\% and Avg = 50\%.

\begin{table}
\begin{center}
\begin{tabular}{ c | c | c | c  }
\hline
  Top 10\% & Top 30\% & Top 50\%  & Avg \\
  \hline
  11 &   29  & 48 & 49
\end{tabular}\label{tabula2}
\caption{Scores in percent for randomly shuffled transition laws.}\label{tabula2}
\end{center}
\end{table}
We introduced the discrete set of states and transition laws. It is very important to see how this system behaves when we add the intermediate states connected with additional laws. In order to mimic the augmented resolution of the physical system,  we only slightly increase the number of the connections of the given state with the another arbitrary state. This allows us to keep the ratio $\varrho = 2.1$ fixed. For the unchanged number of embedding dimensions $d = 10$, the results remain stable up to $10^4$ states. For 1000 states and corresponding 2100 laws we get: Top 10\% = 29\%, Top 30\% = 63\%, Top 50\% = 80\% and Avg = 27\%. The stability of the result is necessary condition if we want to have a good convergence towards the continuous limit and surprisingly, up to the increment of the order $10^3$, the model does not require a higher-dimensional embedding space to keep the results unchanged. This is possibly the consequence of the fact that in $d$-dimensional space, there is $d(d-1)/2$ independent basic rotations (45 in this case where $d=10$), which is obviously enough to align the given distributions. 
\begin{table}
\begin{center}
\begin{tabular}{ c | c | c | c  }
\hline
  Top 10\% & Top 30\% & Top 50\%  & Avg \\
  \hline
  60 &   80  & 95 & 14
\end{tabular}
\caption{Scores in percent for the case when the  outliers are removed.}\label{tabula3}
\end{center}
\end{table}

A significant obstacle in obtaining better results are the outlier states and laws vectors of the embedding (the state 6 in Fig. \ref{graph4}), because they can displace the the whole distribution, which then fails to align properly. When we remove the outliers by hand before orthogonal Procrustes, the results improve significantly (Table \ref{tabula3}). Similar results follow from the translation of the laws vector distribution toward the outlier of the states distribution so that the two distributions match better.

\section{Discussion}\label{sec5}

Our final goal is task-invariant way of acquiring natural knowledge, which is in line with attempts to obtain general (task-independent) artificial intelligence. Here we made first steps towards conceptual generalization on the most basic level. If we have two orders of description of a system, one primitive and one of a higher order, based on the former, we showed one way to put the second-order description into the context of the first-order description. This resulted in unified knowledge representation which could, for instance, allow the learning agent to infer dynamics of the system from kinematical inputs.

One example, where analogous form of learning is present, is the eye-hand coordination at the early stages of human development. Infants take several months to make connections of simultaneous visual, and auditory stimuli with motor activity \cite{Gardenfors2014-GRDTGO, Thelen}. Similarly, humans are able to identify the underlying dynamics of motion, taking into account kinematic information only \cite{Gardenfors2014-GRDTGO, 2002170}. This is in direct correlation with our two-level graph representation, where two types of sequences correspond to kinematics and dynamics of the system. 

Our results suggest that this method or its extensions could be applied to relate two 'languages' of different measure for description of the same physical reality. The co-occurrence of the events on different level amounts to establish causal relations and obtain a clearer picture of the outer world, based solely on the statistics. It refers to any kind of thought representations when there is a possibility to relate objects of cognition with the corresponding meta-objects, based on similar statistics. Instead of the conjectured interpolation, we obtained the triangular structure which might suggest that additional translation has to be applied.

Although this method allows us to establish geometrically based concept formation picture in certain sense, there are several possibilities for refinement. We only used the linear Procrustes transformation. The two distributions of the embedding vectors are only rotated and reflected, but not translated. The analysis from the previous section shows that additional translation remedies the result significantly. This refinement can be achieved by applying additional neural training which could result in translation prior to, or after rotations.  We also tried to obtain alignment using simple generative adversarial network (GAN) \cite{NIPS2014_5423}, but with a limited success. The generator was trained to mimic the distribution of $\bm X$ from the distribution of $\bm Y$. The resulting distribution is similar to the distribution of $\bm X$, but the points of $\bm Y$ tend rather to match and group near to the points of $\bm X$, then to interpolate. Additionally, the stopping criterion should be better understood. One possible way to overcome this problem is the domain-adversarial training \cite{Ganin:2016:DTN:2946645.2946704} which is used in the case of unsupervised machine translation models.

At this stage we did not take into account spatio-temporal relations between the events, but only mutual ordering relation. As we mentioned earlier, up to a slight hint of non-commutativity, there is an open question of ordering too. Namely, the language and thoughts are in general non-commutative (time-ordered) and the word embedding-like representations are basically commutative bags, which are thus not enough for communicating thoughts. Therefore we tried also neural machine translation using Seq2Seq architecture \cite{NIPS2014_5346}, with or without teacher forcing \cite{NIPS2016_6099} or the attention mechanism \cite{DBLP:journals/corr/BahdanauCB14} to order and translate states into laws and conversely. These neural architectures essentially work very well when they are trained on the exact sequences of states and laws. Nevertheless, they only translate without semantic insight, and for the sequences of the states which are not allowed by the given laws, they usually spell nonsensical results. One direction of research would be to train Seq2Seq, not on exact sequences, but rather on environments of given sequences in the embedding space taking into account latent relations of states and transition laws. The other direction could be to apply Memory networks \cite{Weston2015MemoryN,NIPS2015_5846}, with specially prepared datasets or to apply ordering in the form of some sort of positional encoding as in the case of the transformer architecture \cite{NIPS2017_7181}.

The extension of this result can be also interesting in cases of natural supervision \cite{DBLP:journals/corr/abs-1904-12584}, where there is  possibility of simultaneous learning visual concepts, words and sentences. Similarly, it could fit into the framework of the so called the thousand brains theory of intelligence \cite{10.3389/fncir.2018.00121}, which states that brain does not make only one model of the world but many models on different levels, while the resulting concept is the consensus of the votes.

Objects (physical states) and meta-objects (transition laws) in our model are inspired by the concept formation formalism of G\" ardenfors, \cite{Gardenfors2000-GRDCST, Gardenfors2014-GRDTGO}. In this cognitive theory of semantics, notions and meanings can be represented geometrically as convex regions of the so called conceptual space. The underlying idea is to equip the representation space with topological structure, resulting in the existence of the notional domains. The topology is in this context based on convexity, i.e. interpolation (betweenness). Our result may be interpreted from this point of view as follows: if notions $\bm \sigma_a$ and $\bm \sigma_b$ belong to some category (convex domain of space), then the notion $\bm \sigma_a s\,\bm \sigma_b$ which relates notions $\bm \sigma_a$ and $\bm\sigma_b$ belongs to the same notional category. In our case the notion of convexity is replaced by the triangular structure. G\" ardenfors builds a self-contained semantic theory following this assumption, which then explains in-depth the constraints on how words related to their notional dynamical or kinematical nature acquire their syntactic function. This, as well as a neural refinement of the present model, could be possible path for our future research.

\vskip0.5cm\noindent
{\bf Acknowledgement}\  \ 
This work is supported by Kagi.ai Research.

\bibliographystyle{plain}
\bibliography{mybib}

\end{document}